\documentclass[12pt]{article}

\usepackage{scicite}

\usepackage{times}

\usepackage[pdftex]{graphicx}
\usepackage{epstopdf}
\usepackage{amsfonts}
\usepackage{booktabs}
\usepackage{array}
\usepackage{amsmath}
\usepackage{graphics}
\usepackage{epsfig}
\usepackage{multirow}

\usepackage{tabularx}
\usepackage{subfigure}
\usepackage{diagbox}
\usepackage{overpic}
\usepackage{bm}

\newenvironment{sciabstract}{%
	\begin{quote} \bf}
	{\end{quote}}

\title{Biologically Plausible Training of Deep Neural Networks Using a Top-down Credit Assignment Network}

\author
{Jian-Hui Chen$^{1,2,3}$, Cheng-Lin Liu$^{2,3\ast} $, Zuoren Wang$^{1,3,4\ast}$\\
	\\
	\normalsize{$^{1}$Institute of Neuroscience, State Key Laboratory of Neuroscience, }\\
	\normalsize{Center for Excellence in Brain Science and Intelligence Technology, }\\
	\normalsize{Chinese Academy of Sciences, Shanghai, P.R. China}\\
	\normalsize{$^{2}$State Key Laboratory of Multimodal Artificial Intelligence Systems, Institute of Automation,}\\
	\normalsize{Chinese Academy of Sciences, Beijing, P.R. China}\\
	\normalsize{$^{3}$University of Chinese Academy of Sciences, Beijing, P.R. China}\\
	\normalsize{$^{4}$School of Life Science and Technology, ShanghaiTech University, Shanghai, China}\\
	\\
	\normalsize{$^\ast$Corresponding author. E-mail:  zuorenwang@ion.ac.cn, liucl@ia.ac.cn}
}

\date{}

\begin{document}
	
	
	\baselineskip24pt
	
	
	\maketitle

	
\begin{sciabstract}

Despite the widespread adoption of Backpropagation (BP) algorithm-based Deep Neural Networks (DNN), the biological infeasibility of the BP algorithm could potentially limit the evolution of new DNN models. To find a biologically plausible algorithm to replace BP, we focus on the top-down mechanism inherent in the biological brain. Although top-down connections in the biological brain play crucial roles in high-level cognitive functions, their application to neural network learning remains unclear. This study proposes a two-level training framework designed to train a bottom-up network using a Top-Down Credit Assignment Network (TDCA-network). The TDCA-network serves as a substitute for the conventional loss function and the back-propagation algorithm, widely used in neural network training. We further introduce a brain-inspired credit diffusion mechanism, significantly reducing the TDCA-network's parameter complexity, thereby greatly accelerating training without compromising the network's performance.
Our experiments involving non-convex function optimization, supervised learning, and reinforcement learning reveal that a well-trained TDCA-network outperforms back-propagation across various settings. The visualization of the update trajectories in the loss landscape indicates the TDCA-network's ability to bypass local minima where BP-based trajectories typically become trapped. The TDCA-network also excels in multi-task optimization, demonstrating robust generalizability across different datasets in supervised learning and unseen task settings in reinforcement learning. Moreover, the results indicate that the TDCA-network holds promising potential to train neural networks across diverse architectures. In summary, our proposed TDCA-network effectively replaces the traditional loss function and BP training paradigm, and the two-level training framework marks a significant step towards implementing biologically plausible learning in widely used DNNs.

\end{sciabstract}

\section*{Introduction}

Understanding brain intelligence origins is vital in human technology. Artificial neural networks (ANNs), inspired by the biological brain, have made significant progress in artificial intelligence. ANNs derive principles from biological neural networks where neurons form an information processing network. However, ANNs mainly focus on feed-forward information-processing pathways, neglecting the critical top-down connections found in the biological brain.

In neuroscience, top-down connections involve projections from brain areas responsible for higher-order cognitive functions to preceding cortical areas. Research shows that these top-down mechanisms play a key role in higher cognitive functions \cite{kandel2000principles}. For example, top-down projections in the visual system enable a dynamic receptive field, permitting neurons to process relevant perceptual information \cite{gilbert2013top}. In spatial attention, these mechanisms modify specific neuronal populations' activity in the primary visual cortex based on task difficulty \cite{chen2008task}. Additionally, axons in the orbital-frontal cortex offer top-down signals to update coding for inferred decision criteria, crucial for inference-based flexible decision-making \cite{liu2021cortical}. Similarly, projections from the prefrontal cortex to the hippocampus can improve the signal-to-noise ratio for hippocampal encoding of object locations, and augment object-induced increases in spatial information \cite{malik2022top}. Furthermore, several studies have also sought to elucidate the role of top-down projections in credit assignment \cite{stolyarova2018solving,adams2019top,rushworth2011frontal}.
Given the behavioral context information provided by top-down mechanisms, it's essential to explore their introduction into ANNs to expand their functions or improve their performance.

Deep Neural Networks (DNNs) use output errors to update network weight parameters feedback manner. The backpropagation (BP) algorithm, calculating each neuron's synaptic weights update gradients in the network through symbolic chain differentiation, offers a viable solution. This algorithm, widely used for training neural network models, has achieved significant success \cite{rumelhart1986learning}. However, the BP algorithm is not found in the brain \cite{gilbert2013top,tong2003primary}. It requires simultaneous access to all synaptic weights in the network and precise calculation of each weight's gradient, resulting the problems of weight transport \cite{grossberg1987competitive,crick1989recent}, non-local plasticity \cite{baldi2017learning,lowe2019putting}, update locking \cite{czarnecki2017understanding,jaderberg2017decoupled}, and global loss function \cite{marblestone2016toward,journe2022hebbian,richards2019deep, zador2023catalyzing}, which are not biologically plausible \cite{lillicrap2016random,lillicrap2020backpropagation, whittington2019theories, bengio2015towards}. Moreover, the BP algorithm can get trapped in local optima, affecting its optimization ability and the network's representation and generalization abilities \cite{poggio2020theoretical}. Thus, a biologically plausible and efficient algorithm is needed to replace the BP algorithm and further boost DNNs' performance.

During DNN training, the BP algorithm determines the neuron weights that most influence the current task's outcome. This method updates the entire network using a credit assignment mechanism \cite{rumelhart2013backpropagation}. We question if other credit assignment mechanisms could replace the BP algorithm for training DNNs, hence, looking into brain-inspired mechanisms. We present a brain-inspired learning framework that uses a top-down mechanism for credit assignment as shown in Fig. \ref{framework}. Based on the universal approximation theorem, we replace the loss function and backpropagation with a top-down credit assignment neural network (TDCA-network), enabling full network learning without backpropagation or a manually defined symbolic loss function. The network directly generates the parameter gradient. We also establish a credit-diffusion mechanism to further reduce the TDCA-network's parameter complexity, making the framework more efficient and biologically plausible. Additionally, we compare the performance and computational complexity of the neural network-based top-down credit assignment mechanism with traditional back-propagation in different learning paradigms.

\section*{Results} \label{results}

This section applies the TDCA optimization framework to non-convex function optimization, supervised learning, and reinforcement learning. We first compare the TDCA network and the BP algorithm optimizing Gaussian functions, visualizing their gradient fields. In a supervised learning context, we explore the TDCA network's optimization ability across various network structures and datasets, further validating the sparse credit diffusion mechanism's ability to reduce the TDCA network's computational complexity. Finally, in reinforcement learning, we use the TDCA network to train agents for reinforcement tasks of varying difficulty levels, unifying the optimization algorithms for both discrete and continuous action agents.

\subsection*{Application of TDCA Framework on Non-convex Gaussian Function}

To evaluate our top-down framework, we examine the TDCA network's ability to solve non-convex optimization problems with Gaussian functions. We test the TDCA network's multitasking potential by optimizing two different Gaussian functions simultaneously.

\subsubsection*{Non-convex Gaussian function}

\begin{figure}[p]
	\graphicspath{{figures/merge-figures/}}
	\includegraphics[width=\linewidth]{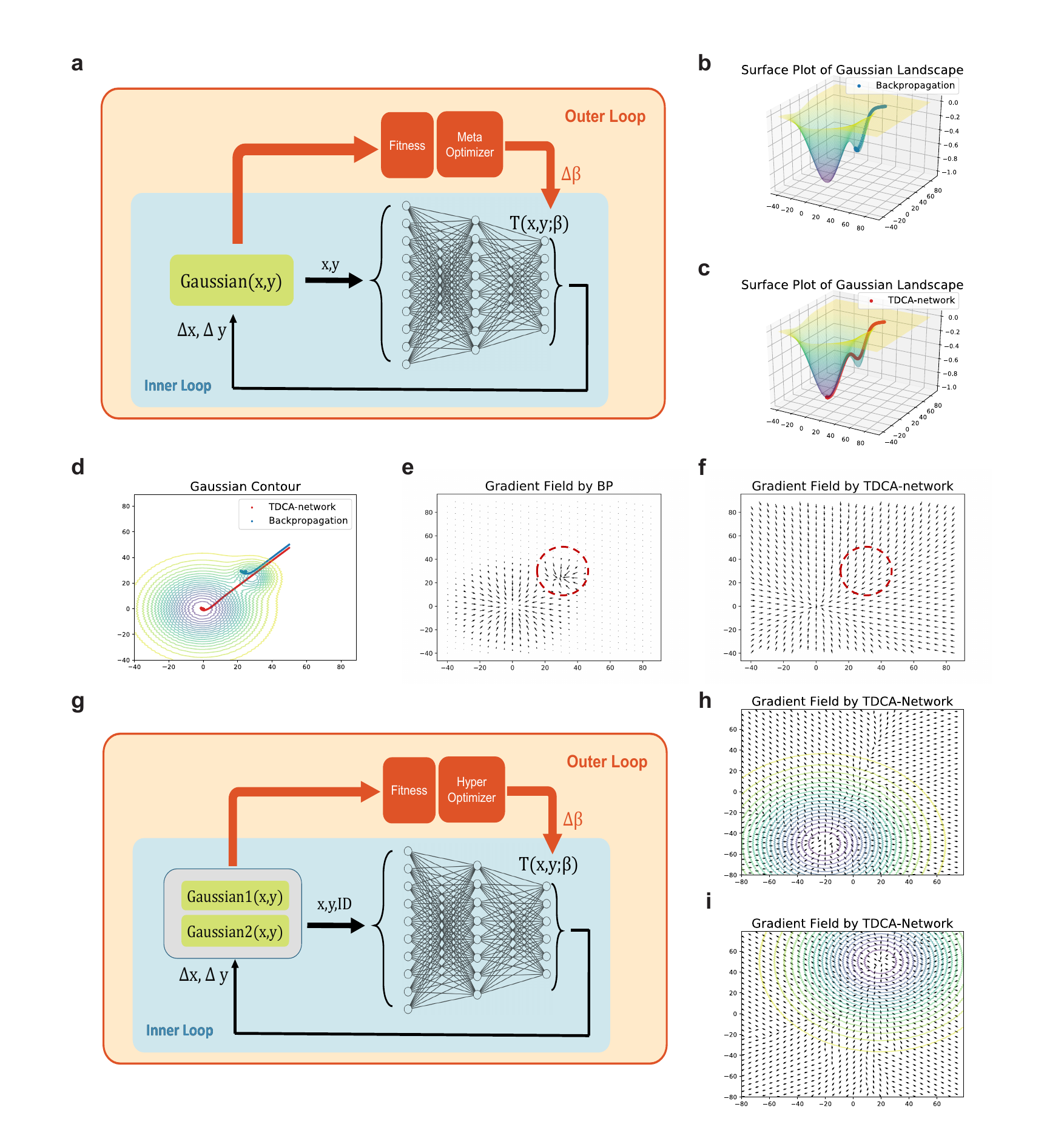}
	
		
		

	\centering
	\caption{Top-Down Learning Framework for Optimization: BP-Free Navigation in Parameter Spaces
	}
	
	\label{convergence_on_gaussian}
\end{figure}

The top-down framework is modified for tow-dimensional function optimization, shown in Fig. \ref{convergence_on_gaussian}(A). We create an optimization task using Gaussian function, consisting of a dominant Gaussian function with a smaller Gaussian function overlay. The optimization parameters are 2-dimensional coordinate values, with a global and a local minimum. We keep the initial points, iteration times, and learning rates identical for different optimization algorithms for easy comparison. The points generated during the optimization process are connected, forming each algorithm's update trajectory.

We compare the update trajectories of the BP algorithm and TDCA network optimization for Gaussian functions. Fig. \ref{convergence_on_gaussian}(B) shows the update trajectory (in blue) of the conventional BP algorithm on the parameter surface. The trajectory stops updating towards the center of the large-scale Gaussian function after reaching the center of the small-scale Gaussian function, indicating the parameters are trapped in a local minimum. Fig. \ref{convergence_on_gaussian}(B) shows the update trajectory by the TDCA network. The TDCA update trajectory moves to the center of the large-scale Gaussian function after crossing the center of the small-scale Gaussian function. This suggests that the TDCA network's optimization strategy can avoid local optimum entrapment, achieving global optimization. Fig. \ref{convergence_on_gaussian}(D) presents the Gaussian contour diagram for both algorithms. A comparison within this diagram shows that the TDCA network can directly reach global optimization, while the BP algorithm is trapped in a local optimum. These findings suggest that the TDCA network algorithm has a superior optimization ability for non-convex Gaussian functions compared to the BP algorithm.

The BP algorithm, a greedy algorithm that only considers local information, updates the coordinates based on the local gradient, making it prone to getting trapped in local optima, despite potentially higher computational efficiency. The TDCA network's update strategy, which cannot be represented using symbolic pseudocode and is not an explicit algorithm, can prevent the optimization process from local optimum entrapment, indicating its superiority over greedy algorithms.

The previous example shows the optimization path from one initialization point. To fully analyze parameter updates guided by the TDCA network, and confirm its optimization capability for non-convex Gaussian functions from any starting point, we calculate the magnitude and direction of the updates for all points on the two-dimensional plane. This establishes a mapping from the coordinate vector to the update gradient vector, also known as a gradient field.

We present the gradient fields of different algorithms on the Gaussian surface. Fig. \ref{convergence_on_gaussian}(E) shows the BP algorithm's update gradient field on the non-convex Gaussian function. Besides gradients pointing towards the global optimum, a group of negative feedback gradients exists at the local optimum (inside the red circle). These gradients trap parameters in this area, leading to their confinement within the local optimum, from which escape is impossible. Thus, the BP algorithm fails to guide towards the global optimal solution. In contrast, Fig. \ref{convergence_on_gaussian}(F) displays the TDCA network's update gradient, free of negative feedback gradients at the red circle. Unlike the BP algorithm, all gradients fom the TDCA network point towards the global optimum.

On much of the Gaussian surface in Fig. \ref{convergence_on_gaussian}(E), gradient magnitudes approach zero, while the gradient in Fig. \ref{convergence_on_gaussian}(F) maintains an adequate intensity. Therefore, the TDCA network's gradients don't saturate. From analyzing gradient direction and magnitude, we infer that, regardless of initial positions, parameters will iteratively update towards the global optimum under TDCA network's guidance, but not with the BP's. The TDCA network's ability to ignore the local optimum and accelerate parameter iteration towards the global optimum confirms its capacity to incorporate global perspectives and offer effective model optimization information.

\subsubsection*{Multi-Gaussian function}

To emphasize the proposed framework's capabilities, we test its optimization of two Gaussian functions at once using a single TDCA network, the framework is shown in Fig. \ref{convergence_on_gaussian}(G). Fig. \ref{convergence_on_gaussian}(H) and Fig. \ref{convergence_on_gaussian}(I) shows contours of two different Gaussian functions, along with the TDCA network's gradient fields at the corresponding positions. We assign Gaussian 1 with task ID vector $[1,0]$ and Gaussian 2 with task ID vector $[0,1]$. The TDCA network generates different gradient fields pointing towards optimal values based on the task ID vector, proving its multitasking optimization ability.

In this subsection, we use the TDCA framework for Gaussian function optimization. Experiments show that the TDCA network's gradient fields can avoid local minima and target the global optimum directly during optimization. Also, the TDCA network can produce separate gradient fields for different tasks, demonstrating its multitasking capabilities. In conclusion, these results affirm the TDCA network's superiority in generating gradient fields for models and forming effective update paths that surpass traditional BP methods.

\subsection*{TDCA Framework Application in Supervised Learning} \label{supervised_learning}

We demonstrate the TDCA optimization framework's utility by examining its supervised learning performance in an image classification task. The experiments include: 1) comparing TDCA network's performance with the BP algorithm for optimizing the classification task; 2) evaluating the sparse credit diffusion mechanism's efficiency in reducing the TDCA network's complexity; 3) studying the convergence process and parameter update trajectory of the bottom-up neural network with both the TDCA framework and the BP algorithm.

We build a fully connected bottom-up network for classification tasks and a fully connected TDCA network for credit signal generation. The TDCA network's ability to train a three-layer bottom-up network with 100 hidden units is assessed using MNIST, Fashion-MNIST, and CIFAR-10 \cite{lecun1998gradient,xiao2017/online,krizhevsky2009learning}. The bottom-up network structure remains consistent, with the exception of varying input numbers across the three datasets: 784-100-10 for MNIST and Fashion-MNIST, and 3048-100-10 for CIFAR-10. The TDCA network uses a 20-30-30-110 neuron configuration. The TDCA network input, generating 110 credits (10 for outputs and 100 for hidden neurons in the bottom-up network), combines the bottom-up network's output with a one-hot label from the dataset. We evolve 5000 generations of the TDCA network parameters using the PGPE algorithm, across all datasets.

\subsubsection*{Test TDCA-network on Different Settings}

We tested the TDCA-network's training efficiency for the bottom-up network using the MNIST dataset under various conditions, such as differing update steps and dataset sizes. We also transferred the TDCA network, trained on the MNIST dataset, to the Fashion-MNIST dataset to evaluate its generalization ability. Table \ref{BP_performance_MNIST} shows the results. The first three rows represent MNIST dataset findings, and the rest outline the Fashion-MNIST dataset results. Each row corresponds to a unique initialization and optimization method for the bottom-up network. We used Kaiming Initialization (``Kaiming BP") and the BP method for network initialization and optimization, respectively. ``Zero BP" represents all bottom-up network parameters initialized to 0 (origin initialization), optimized with the BP method. Owing to the lack of effective gradients at a 0 parameter value, initialization is restricted to within a 1e-16 range around the origin. ``TDCA-net" signifies initializing all network parameters at 0, optimized by the TDCA network. ``TrFC" indicates the transfer of a fully-connected network trained on the MNIST dataset to the Fashion-MNIST dataset. ``TrTDCA-net" denotes the transfer of the TDCA network from the MNIST dataset for training a new bottom-up network on the Fashion-MNIST dataset. The three wider columns represent different dataset sizes and iteration numbers for updating bottom-up networks.

We used the Kaiming initialization method as a positive control due to its widespread use and satisfactory performance. The MNIST dataset results suggest that Zero BP's performance matches the Kaiming Initialization method, indicating that advanced initialization methods may not always excel. However, Zero BP stagnates at a local optimal solution with a training set of 1000 samples. The TDCA network method surpasses local optima, outperforming Zero BP and even the Kaiming Initialization method on smaller training sets.

We tested the TDCA network's generalization capability by transferring it to the Fashion-MNIST dataset. Its performance matches the Zero BP algorithm on training sets of 10,000 and 60,000 samples and surpasses Zero BP on a training set of 1000 samples. Hence, without Fashion-MNIST fine-tuning, the transferred TDCA network from the MNIST dataset (TrTDCA-net) optimizes the bottom-up network (TrTDCA-net vs. Zero BP)effectively. We confirmed this performance by using TrFC as a control. The results reveal that the TrTDCA-net's effect stems from the TDCA network's generalization ability (TrTDCA-net vs. TrFC). In conclusion, the TDCA's update strategy, evolved on the MNIST dataset, can generalize across different datasets.

\subsubsection*{TDCA-network Out-performs BP}

To validate our model, we assessed the TDCA network's optimization of the bottom-up network on multiple datasets: MNIST, Fashion-MNIST, and CIFAR-10. Table.\ref{BP_performance_MNIST} reveals that larger training datasets and more iterations improve performance. Therefore, we set the iteration steps to 150 and utilized the full training dataset. Table. \ref{top-vs-bp-table} compares the performance of TDCA network, Kaiming BP, and Zero BP, indicating that the TDCA network surpasses BP algorithms in fitting and generalization abilities, except for CIFAR-10 training performance. These results highlight the TDCA optimization framework's superiority over traditional loss function and back-propagation methods.

\subsubsection*{Transferability of Well-trained TDCA-network}

We tested the TDCA network's transferability by examining its performance on individual datasets (MNIST, Fashion-MNIST, and CIFAR-10) or combinations thereof. Each TDCA network was evolved using the PGPE algorithm on the mentioned datasets individually and collectively (``ALL"). In the ``ALL" experiment, a single TDCA network optimized three bottom-up networks for each dataset, with the mean cross-entropy across the datasets serving as the PGPE algorithm's fitness evaluation metric. Table. \ref{transfer_table} presents the performance of a TDCA network transferred to different datasets. TDCA networks demonstrate excellent generalization abilities across datasets, regardless of their initial evolution dataset. The training difficulty increases in sequence: MNIST, Fashion-MNIST, CIFAR-10, and ``All", with optimization performance improving with higher dataset difficulty. This trend suggests that evolving TDCA networks on more challenging tasks enhances their generalization ability. Notably, the ``ALL" group shows the best average performance, indicating the advantage of evolving TDCA networks across multiple datasets. Therefore, using more complex, multiple training datasets significantly improves the TDCA network's optimization ability.

\subsubsection*{TDCA-network Performance Across Network Architectures}

We tested the TDCA optimization framework on a deep fully connected neural network (``Deep FC") and a deep convolutional neural network (``Deep CNN") to verify its versatility. The Deep FC has 4 layers with 100 neurons each, and the Deep CNN includes a convolutional layer and three fully connected layers.

Due to the high computational cost of CNNs, we used 10,000 training set examples. Table \ref{top-vs-bp-deep-and-conv} shows the results. The first three rows relate to Deep FC optimization, while the last three concern Deep CNN. Each column shows the dataset results, and each cell presents the classification accuracy and standard deviation for both training and test sets. Comparing these results with Table \ref{top-vs-bp-table} shows that optimizing a 4-layer fully connected network is more complex than a 3-layer one. 

Zero BP with origin initialization underperforms Kaiming BP, highlighting the benefits of Kaiming Initialization. This result suggests that deep fully connected network parameters initialized at the origin tend towards local optima, making the origin a poor initialization point for BP algorithms. However, the ``Top-down" initialized at the origin outperforms Zero BP in all conditions and Kaiming BP on the Fashion-MNIST and CIFAR-10 datasets. In the Deep CNN context, Zero BP trails Kaiming BP in classification accuracy. But, ``Top-down", initialized at the origin, surpasses Zero BP and Kaiming BP on the training set. This finding suggests that Kaiming BP, despite its advanced initialization methods, is still limited by BP algorithm flaws and its final optimization performance is suboptimal. In contrast, the TDCA network performs better, even when initializing deep bottom-up networks at the origin, showing its ability to overcome the local optima challenge. These results highlight the importance of effective optimization algorithms over optimal initialization.

As the network structure complexity increases, optimization becomes more challenging. In these scenarios, the TDCA-net method's optimization effect outshines the Zero BP method, showing its advantage in overcoming local optima. This trend suggests that the TDCA-net method excels in more challenging optimization scenarios, a hypothesis we plan to explore in future studies on more complex optimization problems.

\subsubsection*{Efficient Complexity Reduction of TDCA-network through Credit Diffusion Mechanism} \label{TD-diffuse}

We evaluated the credit diffusion mechanism within the TDCA framework using MNIST, Fashion-MNIST, and CIFAR-10 datasets. The given three-layer network structure constrained us to implement credit diffusion solely on the hidden layer. We defined a neighboring structure for the hidden layer neurons to implement credit diffusion, exploring a one-dimensional line and a two-dimensional grid structure, with each neuron having two and four neighbors, respectively. We set the hidden layer's ``sparsity" - the ratio of credit to neuron count - from 0.1 (10/100, 100/1000) to 0.04 (36/900), using a $\sigma$ of 5 for the Gaussian kernel.

We investigated the impact of credit sparsity and neighboring structures on the bottom-up network's performance. As Table \ref{diffuse_performance} shows, a network of 100 hidden neurons trained with 10 sparse credits (10-to-100) and diffused to adjacent neurons performs similarly to that of 100 dense credits (100-to-100). This suggests that sparse credit diffusion mechanism is effective and can achieve credit sparsity with minimal performance loss.

A 1000-neuron network trained with sparse credit (100-to-1000) outperforms a 100-neuron network trained with dense credit (100-to-100) when the credit count is kept at 100. This indicates that increasing the bottom-up network size while keeping the credit count constant enhances performance. Therefore, the TDCA framework shows promise for controlling large bottom-up network updates using smaller TDCA networks.

A sparsity of 0.04 significantly affects the Fashion-MNIST dataset but has little effect on MNIST and CIFAR-10 datasets (``36-to-900" in Table \ref{diffuse_performance}). Surprisingly, even under extreme credit sparsity, the TDCA framework can complete most bottom-up network-optimization tasks, demonstrating its robust adaptability and indicating that dense gradient information isn't essential for optimization.

We also tested a two-dimensional adjacency structure in the hidden layer, with a 6*6 credit structure controlling a 30*30 neuron structure. The results (in Table \ref{diffuse_performance}) show that the 6*6-to-30*30 performance surpasses both the 36-to-900 performance with identical sparsity, and the 100-to-1000 configuration with denser credit and more neurons. This suggests that higher-dimensional credit diffusion can further enhance the TDCA network's performance.

\subsubsection*{Complexity Analysis} 

Our study quantitatively evaluates the reduction in parameters and computational complexity of the TDCA network using four optimization methods: BP (acting as a control), credit parameters, credit neurons, and credit groups. For bottom-up networks, the credit parameters method often struggles due to the excessive number of parameters. Hence, we provide the TDCA network parameter count for comparison. Table \ref{complexity} summarizes the results, each row corresponding to a unique optimization method, showcasing parameter quantities and computational complexities across various datasets. Computational complexity is expressed as floating point operations (FLOPs).

We analyzed three-layer networks with a hidden layer of either 100 or 900 neurons. By applying our methods to the 100-neuron network, we observed a decrease in the TDCA network's size, parameter quantity, and computational complexity. This suggests that viewing neurons as a whole and using a credit diffusion mechanism can effectively simplify the TDCA network. Moreover, the computational complexity of credit groups rivals the BP algorithm, indicating that replacing BP with our methods won't increase computational complexity. To emphasize the benefits of the TDCA framework, we manipulated the credit's sparsity by using 36 credits to manage the 900 hidden neurons. The table's final two rows show the credit group's computational complexity dropping to nearly one-sixth of the BP algorithm. Thus, when credits are highly sparse, the TDCA network's parameter and computational complexity can be substantially lower than the BP algorithm's. This demonstrates the TDCA network's superior computational efficiency.

\subsection*{TDCA Optimization Framework Applied to Reinforcement Learning}

\begin{figure}[p]
	\graphicspath{{figures/merge-figures/}}
	\includegraphics[width=\linewidth]{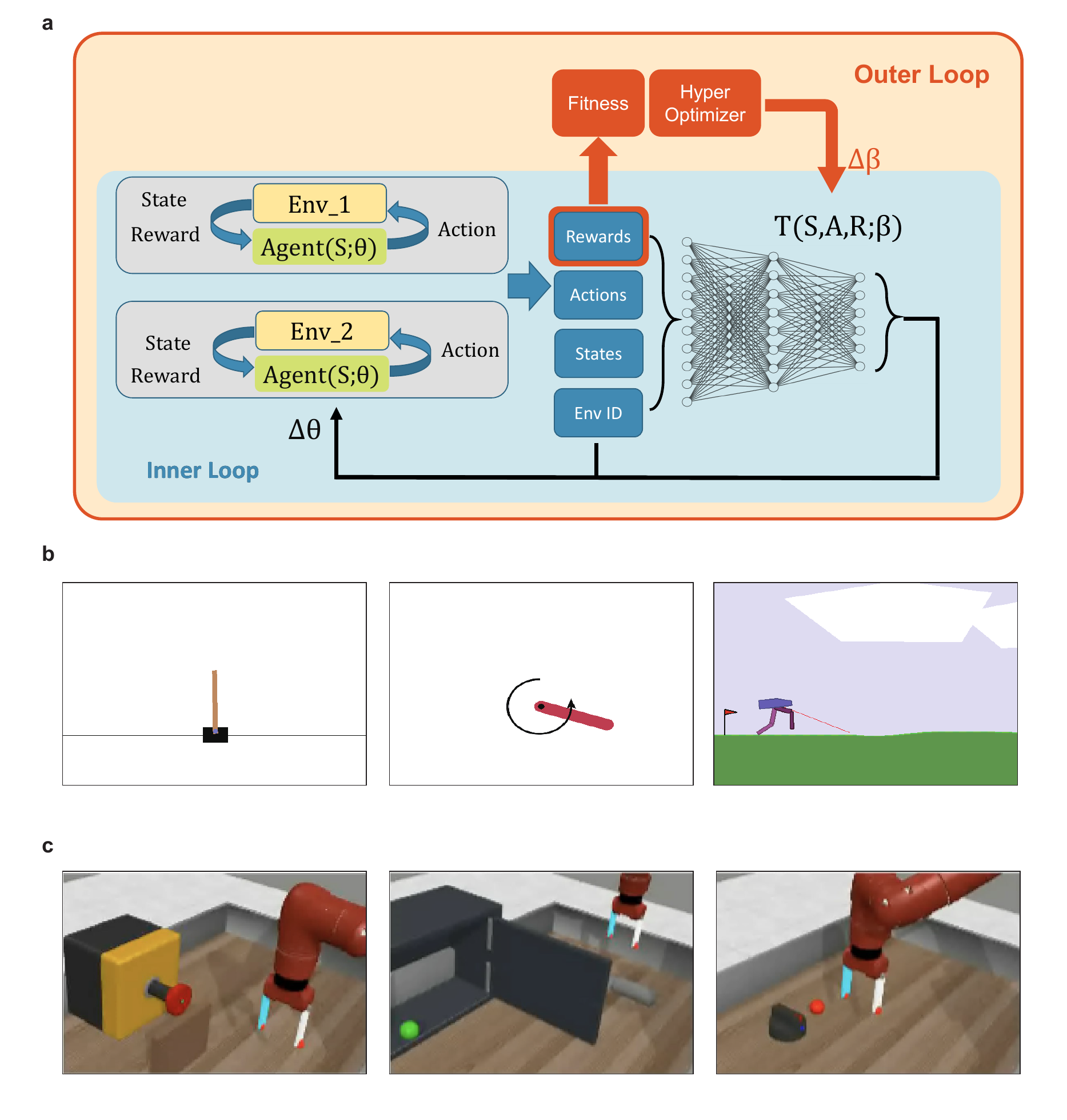}

	
	\centering
	\caption{Multi Reinforcement Task Optimization in Top-Down Learning: Framework and Environments.
	}
	
	\label{reinforce}
\end{figure}

The TDCA optimization framework's adaptability is highlighted through its use in various reinforcement learning tasks. We replace the conventional gradient sampling from agent-generated action outputs and back-propagation with a TDCA network (Fig. \ref{reinforce}(A)).

We conducted experiments in the Gym reinforcement learning simulation environment \cite{1606.01540}, testing the TDCA-framework on CartPole, Pendulum, and BipedalWalker games in ascending difficulty order (Fig. \ref{reinforce}(B)). The TDCA network optimizes agent parameters, aiming to successfully complete the Gym games. An ``agent" in deep learning is a neural network that produces actions for the simulation environment based on the current state, receiving rewards and a new state in return. This interaction generates sequential information, informing our decision to design a RNN structure for the TDCA network. We define the game's duration from start to finish as an ``episode". The TDCA network receives the state, action, and reward vectors from each episode and generates the credit for each parameter, which updates the agent. Given our limited computational resources, we set the agent's update iterations to 200 steps. The TDCA network evolves for 3000 generations under the PGPE algorithm, which has a variance of 0.1, employs the Adam optimizer with a learning rate of 0.1.

Reinforcement learning algorithms fall into two categories: discrete and continuous action. Numerous algorithms cater to these, such as the Policy Gradient (PG) for discrete action, and the Twin Delayed Deep Deterministic Policy Gradient (TD3) for continuous action \cite{silver2014, fujimoto2018addressing}. Table \ref{reinforce-table} compares the performance of the TDCA network with established algorithms. It reveals the TDCA network's success and efficiency in task completion, scoring close to the maxima across various tasks. Two key advantages of the TDCA framework emerge. It optimizes both discrete and continuous action agents, overlooking the action data types difference, indicating its broad applicability. Moreover, it performs on par with the PG algorithm in discrete action tasks and surpasses the TD3 algorithm in continuous action tasks. Despite BipedalWalker's difficulty in continuous action reinforcement learning, the TDCA network completes the task and scores near optimal after only 200 iterations. This efficiency and performance validate the TDCA optimization framework's reinforcement learning abilities.

We tested the TDCA framework's ability to optimize multiple reinforcement learning tasks concurrently using the meta-world environment \cite{yu2019meta}. Fig. \ref{reinforce}(C) illustrates the use of the TDCA framework to optimize agents controlling a robotic arm performing various tasks. With constant degrees of freedom in the robotic arm tasks, we tested agents with identical network structures on different tasks, optimizing various agents with a single TDCA network. We chose three task types for testing: Button-press-wall, Dial-turn, and Door-close. Each task has different initial states, including varying arm initializations and target positions. Tasks with different target positions are considered distinct but share the same paradigm. The TDCA network updates various agents according to the task type ID in its inner loop. We used the average reward score as the fitness metric to estimate the TDCA network updates' magnitude. As shown in Table. \ref{reinforce-multi-task-v2-parallel-table}, the TDCA network's generalization performance was tested on 30 tasks (3 types * 10 initialization) after evolution on 9 tasks (3 task types, each with 3 different initial states). The results show the TDCA network optimized an additional 3*10 tasks effectively without significant score decreases, after training on the 3*3 tasks. This demonstrates the TDCA network's ability to develop distinct optimization strategies per task category, with strategies generalizing within the same category. To explain high reward cases without success (``Botton-press-wall-v2" column), we consulted the Meta-world environment's author, who confirmed this is common.

Our previous experiments on supervised learning showed the TDCA network's generalization capabilities within classification task optimization. And now we show that the TDCA network can achieve generalization within each task type in reinforcement learning and optimize multiple task types concurrently using a single set of hyper-parameters. These results suggest potential practical applications for the TDCA framework.

\section*{Discussion}\label{discussion}

The Backpropagation (BP) algorithm's biological implausibility has prompted significant research efforts for enhancement. Within BP, symmetric weight matrices from symbolic differentiation contradict biological brain characteristics \cite{liao2016important}. Efforts to eliminate this symmetry abound \cite{lillicrap2016random,nokland2016direct,frenkel2019learning,meulemans2021credit}. Feedback Alignment (FA) replaces BP's feedback weight matrix with a random one, supporting error back-propagation in deep learning \cite{lillicrap2016random}. Direct Feedback Alignment (DFA) further streamlines the process by providing feedback from the output layer directly to each hidden neuron, achieving comparable performance to FA \cite{nokland2016direct}. However, FA family, as a BP approximation, retains drawbacks, such as local optima susceptibility, and uses a random feedback matrix with constraints that lack biological plausibility. The BP algorithm also demands global weight information during error back-propagation, driving efforts to localize learning signals \cite{scellier2017equilibrium}. Equilibrium propagation, using an energy-based model to propagate gradient information between neighboring neurons, is one solution \cite{scellier2017equilibrium}. Despite bypassing explicit global information collection, its requirement for local gradient information propagation necessitates more iterations for network learning, slowing adaptation. Likewise, the Difference Target-Propagation (DTP) algorithm back-predicts the current layer's output value based on the next layer's output and updates the weights accordingly \cite{lee2015difference}. Despite avoiding weight symmetry and global information, DTP introduces numerous extra parameters, opposing the biological brain's efficiency. 

Recent advancements in algorithmic design have further enhanced the biological plausibility of optimization algorithms \cite{hinton2022forward, dellaferrera2022error, srinivasan2023forward, journe2022hebbian, meulemans2021credit, song2024inferring}. However, these studies predominately concentrate on the back-propagation aspect within neural networks. Our research approaches the loss function and back-propagation as an integrated entity, addressing their biological plausibility in a comprehensive manner.

In this study, we introduce an innovative training framework for optimizing neural networks, inspired by the top-down projection architecture found in biological brains. Our framework employs the Top-Down credit Assignment (TDCA) neural network to supplant the conventional back-propagation (BP) algorithm in enhancing bottom-up neural networks. We have validated the efficacy of the TDCA network in supplanting the traditional optimization paradigm which characterized by loss functions and gradient descent across different optimization settings, such as non-convex function optimization, supervised learning, and reinforcement learning. Our empirical investigations reveal that TDCA network-infused optimization strategies circumvent local optima entrapment, a common hindrance in BP-based parameter tuning. By leveraging a neuroparacrine-inspired sparse credit assignment mechanism, the TDCA-network outperforms the BP algorithm in terms of optimization results, while requiring fewer computational resources. This innovative approach, wherein one neural network optimizes another, provides valuable insights for design philosophies in the field.

Conventionally, output derived from bottom-up models undergoes integration with environmental signals to assess final performance via a symbolic loss function, followed by parameter adjustments through gradient descent. This process essentially maps the model output and environmental data to model parameter update gradients, with the mapping's specifics dictated by the designed loss function and gradient application, thereby defining the parameter optimization strategy. Our training framework automates this traditionally human-designed segment by utilizing evolutionary algorithms to discover novel optimization strategies. 
Prior attempts at brain-like models, including feedback alignment, equilibrium propagation, and difference target-propagation algorithms, have strived to resolve the dissonance between symbolic computation and the network-based implementation of brain functions. Crucially, even if these algorithms resolve gradient propagation issues, their gradients still originate from loss function derivatives. In contrast, our framework tackles the issue at its root, comprising solely neural networks for optimization, akin to the brain's methodology.

Our framework is exceptionally robust, unifying diverse learning algorithms and exhibiting formidable optimization prowess across various application domains. Nevertheless, it presents open questions, such as the extent to which the TDCA network's sparse contribution diffusion mechanism can decouple its parameter count from that of the optimized bottom-up model, its effectiveness in optimizing large neural structures, the number of tasks it can concurrently optimize, its applicability to unsupervised learning, and its performance across different task scenarios. A critical consideration is the reliance on evolutionary algorithms for TDCA optimization, which is computationally intensive. Hence, more efficient optimization strategy search algorithms are imperative for broader application of our framework.

Since the interconnectedness of machine learning and neuroscience is well-established, our optimization framework signifies a paradigm shift in machine learning, wherein neural networks is the searching space of optimization algorithm strategies, and evolutionary algorithms facilitate the discovery of innovative strategies beyond the traditional, manually-designed objective functions and gradient back-propagation. Concurrently, in neuroscience, this fully networked framework suggests that top-down projections in the brain may be integral to learning processes, offering fresh insights into brain functionality. Therefore, our contribution holds promise for substantial advancements in both machine learning and neuroscience.

\section*{Acknowledgements}\label{sec:Acknowledgments}
This work has been supported by the STI2030-Major Projects (2022ZD0205100) to ZW, and the National Natural Science Foundation of China (NSFC) grants 61836014 to CL.

\clearpage
\section*{Materials and Methods}\label{methods}

We present a two-level learning framework, guided by brain-like top-down mechanisms, using a TDCA-network to replace traditional loss functions and back-propagation processes. Our work breaks down into four parts: 1) TDCA-network's replacement of back-propagation and optimization; 2) non-convex function optimization adaptation; 3) supervised learning adaptation; and 4) reinforcement learning adaptation.

\subsection*{The Top-down Credit Assignment Framework}

Our goal is to create a fully networked optimization framework, consistent with brain-like information flow. The bottom-up network performs environment-driven tasks. Typically, back-propagation, steered by a loss function, produces this network's parameter gradients. We suggest a top-down network that replaces both the loss function and back-propagation to directly generate gradients. This top-down network, responsible for credit assignment, receives information from top-level representations to bottom-level inputs.

\subsubsection*{Overview of the Framework}
\begin{figure}[p]
	\graphicspath{{figures/}}
	\centering
	\includegraphics[width=\linewidth]{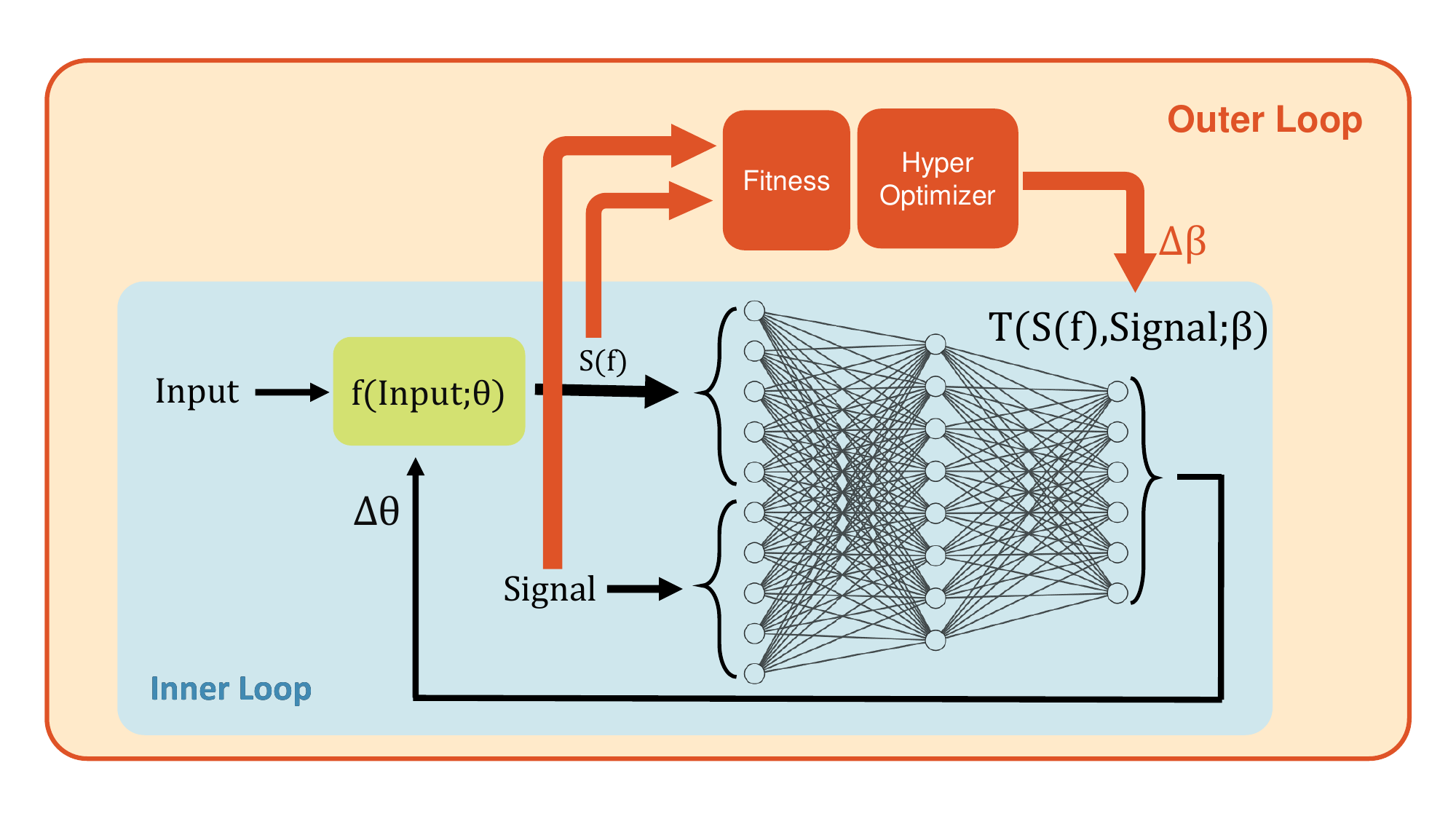}
	\caption{The schematic diagram of our novel top-down learning framework. 
	}
	\label{framework}
\end{figure}

Fig.\ref{framework} depicts the framework with two loops: an inner loop optimizing the bottom-up network and an outer loop working on the TDCA network.

In the inner loop, the bottom-up network $f(Input;\theta)$, controlled by parameter $\theta$, produces outputs based on $Input$. The TDCA-network then generates parameter gradients $\Delta \theta$ for the bottom-up network, using state $S(f(Input;\theta))$ (or $S(f)$) and the environment $Signal$. We explore the state $S(f)$ definition in later sections. The TDCA-network as function $T(S(f),Signal;\beta)$ uses parameter $\beta$ to control the credit-assignment strategy.

The outer loop, shown in orange in Fig. \ref{framework}, optimizes parameter $\beta$ using a hyper-optimizer, based on measurements from $S(f)$ and $Signal$. The bottom-up network is re-initialized at the beginning of each inner loop; the final state $S(f)$ transmits to the outer loop after each run. The TDCA-network outputs continuously update the parameter $\theta$ of the inner loop's bottom-up network, leading to complex dynamics. We use a black-box optimization algorithm as the hyper-optimizer to circumvent differentiating hyper-parameter $\beta$ in this dynamic system. The Evolution Strategy, a black-box optimization technique, has demonstrated success in optimizing large-scale neural networks, matching the performance of leading optimization algorithms \cite{salimans2017evolution}. The update rule under PEPG are detailed in supplementary.

The evolution strategy algorithm, despite matching state-of-the-art results, has a higher computational cost \cite{zhang2017relationship}. Two key points need attention. First, PGPE can handle discrete fitness functions, but continuous functions provide richer information and require fewer sampling points. Second, as the number of parameters increases, so does the optimization difficulty, sampling points, and computational power. To enhance practicality and reduce power requirements, we adjust the TDCA optimization framework by reducing the TDCA network parameters. The optimization process is detailed in the following sub-sections.

\subsection*{Non-convex Function Optimization}


We first apply the TDCA optimization framework to simple problems. We adjust the top-down learning framework (see Fig. \ref{convergence_on_gaussian}(A)) to use the TDCA network in a two-dimensional Gaussian function optimization. Here, the inner loop's optimization objective is the Gaussian function $Gaussian(x,y)$, with $x$ and $y$ coordinates representing the optimization parameters. The Gaussian function's mean and variance are fixed. Without additional input, the Gaussian function's state is the current coordinate pair $[x, y]$. The TDCA network generates the corresponding gradients $\Delta x$ and $\Delta y$ and seeks the Gaussian functions' extremum points. Concurrently, the outer loop's hyper-optimizer updates the parameter $\beta$ based on the Gaussian function's value after each inner loop run.


The TDCA framework can also optimize two distinct Gaussian functions. We adapt the framework to optimize multiple Gaussian functions simultaneously (see Fig. \ref{convergence_on_gaussian}(G)). We treat two Gaussian functions with different means and variances as separate tasks, each with a unique one-hot ID. When the TDCA network optimizes a Gaussian function, we input the task ID with the current coordinate values. The TDCA network generates different gradients for the same coordinates based on the task ID. This process can be challenging and needs more iterations for the TDCA network to understand the task ID's meanings. Thus, in subsequent experiments, we adjust the network structure to account for the correlation between task IDs and output gradients, decreasing the optimization process's computational cost.

\subsection*{Supervised Learning}


We use the TDCA framework to optimize feed-forward networks for image classification tasks, shown in Figure. \ref{diffusion}(A). The feed-forward (bottom-up) network $f(x;\theta)$ predicts the class label probability distribution $p(\hat{y}|x)$ for an input image $x$. The TDCA network calculates the parameter gradient from the feed-forward network's state $S(f)$ and label $y$. The network state $S(f)$ is defined using back-propagation, which we can express as

\begin{equation}
\Delta \theta = Backpropagation(Loss(f(x;\theta),y), x, \theta).
\end{equation}

Replacing the loss function and back-propagation process, the only feed-forward network variables are input $x$ and parameters $\theta$. Thus, we define the feed-forward network state, $S(f)=[x, \theta]$. We discuss how this vector projects to $p(\hat{y}|x)$ in sub-section \ref{reduce_input_dim}. Given the current state $S(f)$ and label $y$, the TDCA network's credit assignment strategy is determined by parameter $\beta$.

For large-scale feed-forward network optimization, we need to reduce computational complexity and improve efficiency. In classification tasks, we use cross-entropy as the fitness function for the outer loop to decrease the necessary sample number. By controlling the TDCA network size through input and output dimension reduction, we further reduce the number of hyper-parameters the outer loop optimizes.

\subsubsection*{Reducing TDCA-network Input Dimension} \label{reduce_input_dim}

The TDCA-network input is the feed-forward network state, and its dimension depends on the state $S(f)$. If we define the state using the full dimension of $x$ and $\theta$, the TDCA-network becomes too large. Thus, we need to define the state effectively to include as much feed-forward network information as possible while keeping low dimensionality. We can reduce the dimensions of $x$ and $\theta$. Conveniently, the feed-forward network output $f(x;\theta)$ reduces dimensionality by projecting two parameters to each class's probability distribution. Hence, we can use the feed-forward network output as the TDCA network input, significantly reducing its input dimension.

\subsubsection*{Reducing TDCA-network Output Dimension}


We aim to reduce the output dimension of the TDCA network, closely related to the feed-forward network. For a single-layer network with $n$ neurons, each receiving an $m$-dimensional vector, the network comprises $n * m$ parameters. If each parameter is controlled by the TDCA network output, the output dimension equals $n * m$. However, by assigning a single credit to adjust all $m$ parameters of each neuron, we diminish the TDCA network output dimension to $n$. This reduction is realized through an internal credit distribution mechanism (see Figure. \ref{diffusion}(B)), where each neuron executes local Hebbian learning based on the input and provided credit signal. Consequently, we reduce the TDCA network output dimension by a factor of $m$.

\subsubsection*{Sparse Credit Assignment in Top-down Framework}

We've conceptualized neurons as unified entities to trim the TDCA network's output dimension, greatly reducing its parameter count. However, large bottom-up networks still require substantial TDCA network output for credit allocation. Therefore, a mechanism to further shrink the output dimension is essential, especially for large networks.

Neuroscience provides evidence of synapse crosstalk, and corresponding computational models exist \cite{barbour1997intersynaptic}. Critical neurotransmitters in the central nervous system, Glutamate and GABA, can function extra-synaptically \cite{sem2005diffusional}. These neurotransmitters, when released, can diffuse and modulate nearby neurons. Dopamine and serotonin, neurotransmitters vital for learning and memory, contact multiple synapses extensively, influencing proximate neurons \cite{liu2019mechanisms,ogren2008role}. Hence, a neuron's neurotransmitters commonly modulate neighboring neurons in a paracrine manner.

Borrowing from these paracrine mechanisms, we've integrated a credit diffusion mechanism into the TDCA framework. We've defined a neighboring structure to emulate signal spread among neurons within the same layer. As shown in Fig.\ref{diffusion}(C), credits are initially sparsely allocated to neuron subsets, unlike the dense assignment in Fig. \ref{diffusion}(D), and then diffused across the entire layer. A pre-set Gaussian kernel function controls the attenuation process, reflecting distance variations between neurons during diffusion. As neuron distance expands, the credit impact decreases. For a credit $C$ assigned to neuron $i$, its diffusion to neuron $j$ within the same group is:

\begin{equation}
C*e^{\frac{-(Coord(i)-Coord(j))^2}{\sigma^2}}, 
\end{equation}
where $Coord(\dot)$ returns the neuron structure's coordinates and $\sigma$ sets the credit diffusion distance.

This method allows us to update a group of bottom-up neurons with a single credit, reducing the credit count needed to update the bottom-up network. This facilitates more efficient bottom-up model training with a smaller TDCA network. If a group contains $k$ neurons, the TDCA network's credit dimension further reduces by a factor of $k$. Thus, the credit diffusion mechanism significantly diminishes the TDCA network's complexity. We can either use a smaller TDCA network with the same bottom-up network structure, or train larger bottom-up networks with the existing TDCA network structure for improved performance.

%

\subsection*{Reinforcement Learning}

\subsubsection*{Optimizing Single Reinforcement Task}


To further investigate the TDCA optimization framework's applicability, we apply it to reinforcement learning. Figure \ref{reinforce}(A) illustrates a bottom-up network, functioning as an agent, interacting directly with the reinforcement learning environment. The agent produces actions from the current state of the environment, leading to state updates and corresponding rewards. The environment resets before each interaction and reaches a terminal state after a predetermined number of interactions, which we define as an episode. After each episode, the agent's parameters change to maximize future rewards using the TDCA network. This network generates the parameter update gradients $\Delta \theta$, using information from the current episode like environment states, actions, and rewards. Given the sequence input, the TDCA network uses a recurrent neural network structure. We expect the agent to adapt to its environment through iterative inner loop processes, achieving maximum reward within an episode. We evaluate the TDCA framework's ability to optimize multiple reinforcement tasks. When one TDCA network is used to optimizes multiple agents performing reinforcement learning tasks, we assign unique IDs for each  task, allowing the TDCA network to generate update gradients for each task based on its ID.

In the outer loop, we collect the final iteration's cumulative reward or the average reward of multiple tasks to evaluate the current TDCA network. The outer loop's PGPE optimization algorithm estimates $\Delta \beta$ from actual samples and updates the TDCA network. With the bottom-up network's small size, additional methods to reduce the TDCA network's parameter count are not needed.

\bibliographystyle{Science}
\bibliography{Thesis}

\clearpage

\section*{Figure Legends}
\subsection*{Figure.\ref{framework} The schematic diagram of our novel top-down learning framework.}

There are two neural networks in the framework. One is the bottom-up network, which is responsible for making appropriate judgments and actions based on the input from the environment; the other is the top-down credit assignment (TDCA) network, which optimizes the bottom-up network by integrating the internal information $S(f)$ from the bottom-up network and the external information $Signal$ from environment. The framework consists of two parts: the inner loop (blue) and the outer loop (orange). (1) The inner loop uses the optimization strategy of the TDCA-network to optimize the parameter $theta$ of the bottom-up network. (2) The outer loop uses an evolutionary algorithm to optimize the parameters $\beta$ of the TDCA-network. In the current scenario $\beta$ is a hyper-parameter, which controls the optimization strategy for the bottom-up network. By adjusting the parameter $\beta$ multiple times, we will obtain a TDCA-network corresponding to an effective optimization strategy.

\subsection*{Figure.\ref{convergence_on_gaussian} Top-Down Learning Framework for Optimization: BP-Free Navigation in Parameter Spaces}

\textbf{a}, We adapted the top-down learning framework for mixed Gaussian optimization. Coordinate parameters to be optimized, $x$ and $y$, are feed into the TDCA-network, which then produces a gradient used to update these parameters. Simultaneously, the TDCA-network is optimized to generate improved gradients for each coordinate within the Gaussian parameter space.		

\noindent 
\textbf{b}, The blue-labeled trajectory represents the bottom-up model optimization trajectory trained by BP, which falls into a local minima. 

\noindent  
\textbf{c}, The red-labeled trajectory represents the TDCA-network, which navigates directly to the global minima, achieving the global optimal solution. 

\noindent 
\textbf{d} This panel compares the update trajectories of bottom-up networks trained by BP (blue) and the TDCA-network (red) within the 2-dimensional Gaussian contour.

\noindent 
\textbf{e}, Gradient field under BP-based optimization, gradients are oriented towards both local and global minima. 

\noindent 
\textbf{f}, TDCA-network-based Gradient field, gradients lead directly to the global minimum, disregarding the local minimum.

\noindent 
\textbf{g}, Multiple Gaussian function-optimization tasks, featuring a single TDCA network and two distinct Gaussian functions. Depending on the task number ID, the TDCA network can independently optimize the parameters of Gaussian1 and Gaussian2. Fitness is the average value of the two Gaussian functions.

\noindent 
\textbf{h} and \textbf{i} present contour maps of two distinct Gaussian functions and the corresponding gradient field generated by the TDCA network for two distinct Gaussian functions. It is apparent that all gradient points to the global optimum of their respective tasks.

\subsection*{Figure.\ref{reinforce} Multi Reinforcement Task Optimization in Top-Down Learning: Framework and Environments.}

\textbf{a}, Schematic diagram of multi reinforcement task optimization in top-down learning. The framework comprises two agents (each corresponding to different environments) realized by a neural network and a shared TDCA network. The agents interact with their respective environments, accumulating rewards until the tasks conclude. The TDCA network generates the gradients to update parameters $\theta$ for both agents, distinguishing the gradients for each agent by the environment ID. In the outer loop, the hyperparameter optimizer evaluates the performance of the TDCA network by averaging the rewards of the two agents after each inner loop run, then adjusts the hyperparameter $\beta$.

\noindent 
\textbf{b}, The reinforcement learning environment of the Gym platform. We evaluated three Gym reinforcement environments of varying difficulty levels (from easy to difficult): cart pole, pendulum, and bipedal walker. The cart pole game employs a discrete action space, while the pendulum and bipedal walker utilize continuous action spaces.

\noindent  
\textbf{c}, The reinforcement learning environment of the Meta-World platform. We assess three types of reinforcement tasks related to robotic arm control: Button-press-wall, Door-close, Dial-turn.

\begin{figure}[p]
	\graphicspath{{figures/merge-figures/}}
	\centering
    \includegraphics[width=0.7\linewidth]{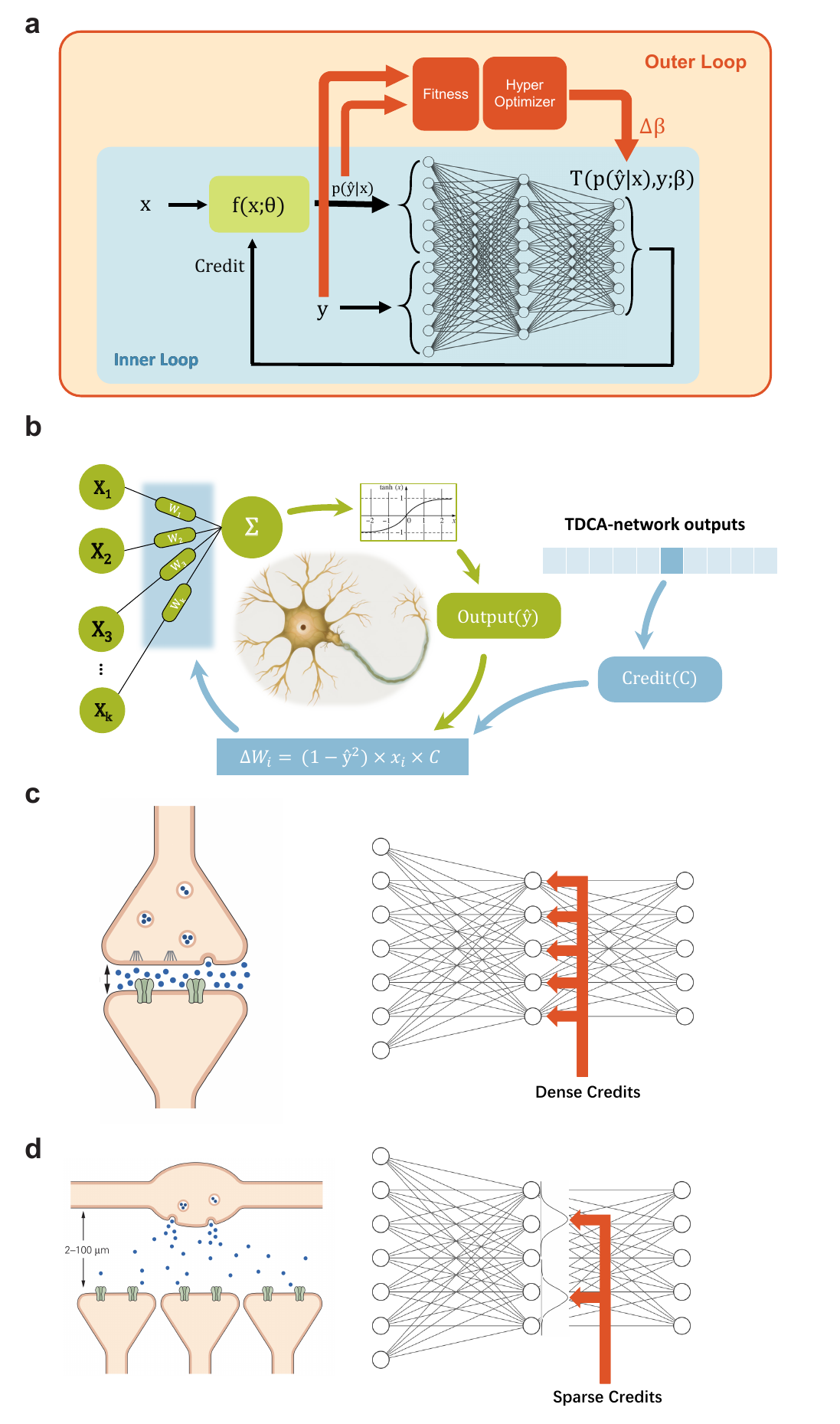}
	

	\caption{Top-Down Credit Assignment Learning Framework for Classification Tasks in Supervised Learning: Dense and Sparse Credit Distribution.
	}
	\label{diffusion}
\end{figure}

\subsection*{Figure.\ref{diffusion} Top-Down Credit Assignment Learning Framework for Classification Tasks in Supervised Learning: Dense and Sparse Credit Distribution}

\textbf{a}, The top-down credit assignment learning framework for classification tasks in supervised learning. The framework uses an image $x$ as the input for the feed-forward network, which in turn generates probabilities attributing the image to each category. The TDCA network generates a credit based on the output probabilities $p(\hat{y}|x)$ and the image category label $y$, which is used for parameter updates. In the outer loop, the $\beta$ hyperparameter optimizer evaluates and updates the TDCA network's optimization strategy based on the classification accuracy or cross-entropy after each inner loop run.

\noindent 
\textbf{b}, Upon receiving credit $C$ from the TDCA network, each neuron's weight is updated according to the Hebbian learning rule. The update gradient for each weight is calculated by multiplying the corresponding input $x_i$ and Credit $C$. Given that our neurons use Tanh as the activation function, the gradient incorporates Tanh's derivative as a coefficient. The synaptic weights are then updated according to these gradients.

\noindent 
\textbf{c} Schematic diagram of dense credit, with each neuron obtaining one credit value. 

\noindent  
\textbf{d} Sparse credit is applied to a layer of neurons, with groups of neurons sharing one credit value. This value decreases with distance, mimicking natural diffusion. Figures of synapse are adapted form \cite{kandel2000principles}

\clearpage

\clearpage

\clearpage

\clearpage

\clearpage

\begin{table*}
	\centering
	\caption{Classification performance under different optimization methods and settings.}
	\resizebox{\linewidth}{!}{
		\renewcommand{\arraystretch}{1.25}
		\begin{tabular}{cccccccc}
			\hline
			\multirow{2}{*}{Dataset}      &   \multirow{2}{*}{Method} &  \multicolumn{2}{c}{20 step} & \multicolumn{2}{c}{50 step} & \multicolumn{2}{c}{150 step}   \\
			\cmidrule(lr){3-4}  \cmidrule(lr){5-6}  \cmidrule(lr){7-8}
			& & Train(1000) & Test(10000) & Train(10000) & Test(10000) & Train(60000) & Test(10000) \\ 
			\hline 
			\multirow{3}{*}{MNIST} &   Kaiming BP     &  $96.00\pm0.88\%$  & $88.14\pm0.23\%$ & $ 96.02\pm0.095\% $& $93.12\pm0.16\%$ & $98.20\pm0.15\%$ & $96.16\pm0.14\%$ \\
			&   Zero BP    & $90.96\pm 0.33 \%$& $85.31\pm0.28\%$ & $ 97.23\pm0.093\% $& $93.43\pm0.083\%$ & $99.04\pm0.07\%$ & $96.35\pm0.15\%$ \\
			&   TDCA-net	& 99.8\% & 88.61\% & 96.54\% & 93.56\% & 99.08\% & 96.82\% \\
			\hline
			\multirow{4}{*}{Fashion-MNIST}   & Kaiming BP     &  $84.28\pm2.34\%$  & $76.06\pm1.25\%$ & $87.592\pm0.50\%$ & $83.63\pm0.26\%$ & $91.26\pm0.37\%$ &$ 87.19\pm0.41\% $\\
			& Zero BP     &  $72.5\pm0.59\%$  & $67.43\pm0.81\%$ & $87.37\pm0.14\%$ & $83.47\pm0.25\%$ & $92.33\pm0.11\%$ &$ 87.48\pm0.12\% $\\
			& TrFC         &  8.2\% & 10.00\% & 8.73\%  & 9.31\%  & 5.38\%  & 5.18\% \\
			& TrTDCA-net	& 76.1\% & 73.29\% & 78.28\% & 76.99\% & 85.94\% & 83.91\% \\
			\hline
			
	\end{tabular}}
	\label{BP_performance_MNIST}
\end{table*}

\begin{table*}
	\centering
	\caption{Classification performance on different network structure and datasets. }	
	\resizebox{\linewidth}{!}{
		\renewcommand{\arraystretch}{1.25}
		\begin{tabular}{ccccc}
			\hline
			Model & Method &  MNIST  & Fashion-MNIST  & CIFAR-10  \\
			\hline
			\multirow{3}{*}{Deep FC} & Kaiming BP   &  $99.83\pm0.05\% / 96.78\pm0.20\% $ & $91.98\pm0.66\% / 86.75\pm0.49\% $& $58.17\pm0.99\% / 38.07\pm0.63\%$ \\
			
			& Zero BP                &  $ 98.97\pm0.15\% / 95.96\pm0.21\%$ & $90.51\pm0.58\% / 86.00\pm0.39\%$ & $ 48.34\pm0.87 / 38.19\pm0.57 $ \\	
			& TDCA-net                     &  99.06\% / 96.71\%  & 93.27\% / 87.71\% & 58.76\% / 41.70\% 	\\
			\hline
			
			\multirow{3}{*}{Deep CNN (10000)} & Kaiming BP  &  $99.98\pm0.009\% / 95.86\pm0.28\% $ & $93.90\pm1.12\% / 84.87\pm0.09\% $& $75.13\pm2.11\% / 44.13\pm1.29\%$ \\
			& Zero BP                &  $ 95.79\pm0.45\% / 94.34\pm0.33\%$ & $87.09\pm1.31\% / 83.54\pm0.72\%$ & $ 34.97\pm2.45\% / 36.20\pm2.82\% $ \\	
			& TDCA-net                     &  99.97\% / 95.61\%  & 98.45\% / 82.54\% & 90.66\% / 34.98\% 	\\
			\hline

	\end{tabular}}
	
	\label{top-vs-bp-deep-and-conv}
\end{table*}

\clearpage
{	
\centering
\section*{Supplementary Information}
}

\indent\textbf{~~~~~~~~~~1.Supplementary Methods and Results}

\textbf{~~~~~~~~~~2.Figure S1-S3}

\textbf{~~~~~~~~~~3.Table S1-S6}

\setcounter{figure}{0}
\renewcommand\thefigure{S\arabic{figure}}

\setcounter{table}{0}
\renewcommand\thetable{S\arabic{table}}

\clearpage

\subsection*{Evolving Parameters of The TDCA-network}
We apply the Policy Gradients with Parameter-based Exploration (PGPE) algorithm in this study to approximate gradient calculations via sample fitness \cite{sehnke2010parameter}. The PGPE algorithm optimizes $\beta$ in our framework through three steps: 1) Drawing samples $\hat{s}$, using $\beta$ as the sampling center, from a Gaussian distribution; 2) Training the bottom-up network with each sampled hyper-parameter $\hat{s}$ and assigning fitness based on training results; 3) Estimating the $\beta$ gradient based on overall sample fitness.

Regarding the PGPE algorithm, we define the Gaussian distribution from which we sample as:

\begin{equation} \label{guassian}
p_{\beta}(s)=\mathcal N(\beta, \sigma I); ~s = \beta + \sigma \mathcal \epsilon, \epsilon \sim \mathcal N(0,I),
\end{equation} 
where $\sigma$ is the sampling variance. The PGPE algorithm draws $N$ samples $\hat{s}$ from the distribution $p_{\beta}(s)$ to estimate the gradient expectation at $\beta$.

We use the fitness function $F(\hat{s})$, based on the sample performance. In a classification task, for instance, gradients are generated for the bottom-up network to improve classification performance. Hence, the fitness function is defined as:

\begin{equation}
F(s) = Accuracy(f(x;\theta_{M}), y), 
\end{equation} 
where subscript $M$ denotes the iteration times of inner loop, and

\begin{equation} 
\begin{split}
\theta_{i+1} = \theta_{i} + T(S(f(Input;\theta_{i})),y;s),  i \in 1,2,3...,M
\end{split}
\end{equation} 

Taken together, the expectation of fitness on $\beta$ is 

\begin{equation} \label{fitness}
\mathcal F(\beta) = \mathbb{E}_{s \sim p_{\beta}(s)} F(s)= \int_{s} F(s)p_{\beta}(s) ds.
\end{equation} 

Finally, the gradient of $\beta$ is (please refer to \cite{salimans2017evolution} for more details)

\begin{equation} \label{gradient}
\begin{split}
\nabla_{\beta}\mathcal F(\beta) = \nabla_{\beta} \mathbb{E}_{s \sim \mathcal N(\beta, \sigma I)} F(s)\\
=\nabla_{\beta} \mathbb{E}_{\epsilon \sim \mathcal N(0,I)} F(\beta + \sigma \epsilon)\\
=\frac{1} {\sigma} \mathbb{E}_{\epsilon \sim \mathcal N(0,I)} {\epsilon F(\beta + \sigma \epsilon) }.
\end{split}
\end{equation} 

\subsection*{Experiments on PGPE}

\begin{figure}[p]
	\graphicspath{{figures/analysis/}}
	\subfigure[Cross-Entropy]{
		\includegraphics[width=0.45\linewidth]{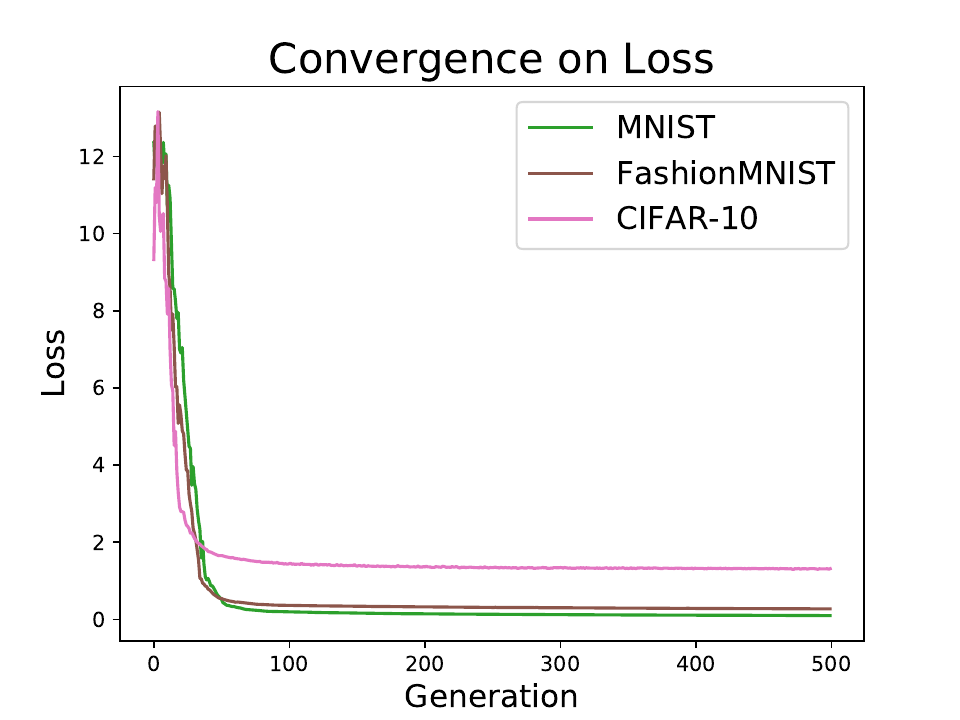}
		\label{PGPE_CE}
	}
	\subfigure[Accuracy]{
		\includegraphics[width=0.45\linewidth]{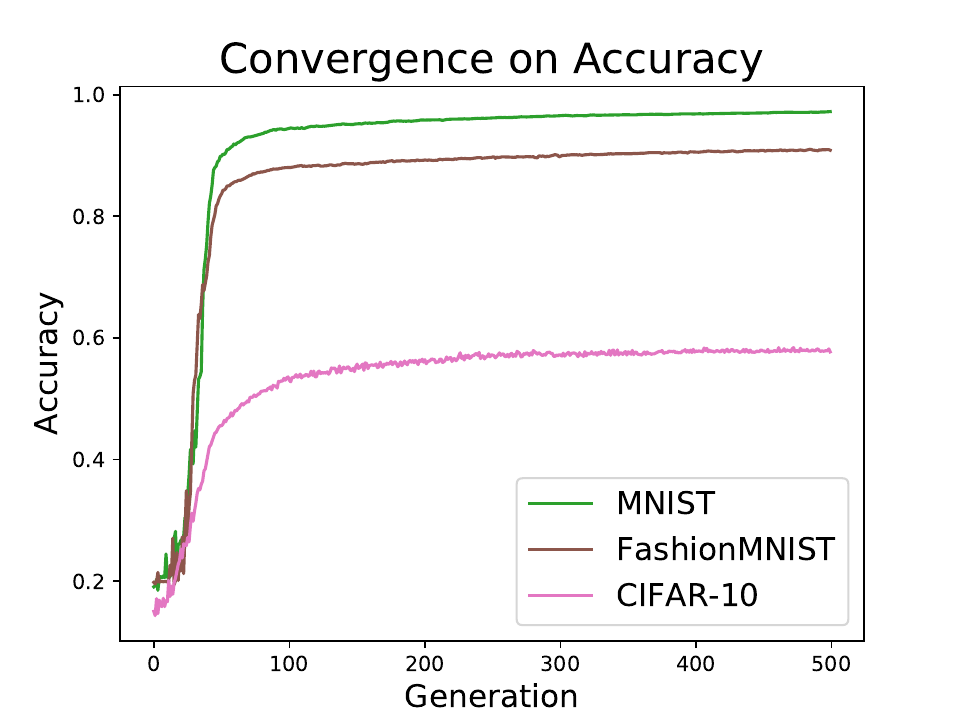}
		\label{PGPE_acc}
	}
	
	\centering
	\caption{Convergence of the PGPE algorithm during TDCA-network training. The first 500 iterations of the TDCA network training's outer loop are selected for analysis. Convergence curves of these two metrics across various data sets are depicted in the figures. Panels (a) and (b) illustrate the convergence process of cross-entropy and accuracy, respectively.}
	\label{PGPE_convergence}
\end{figure}

We tested the efficacy of the PGPE algorithm in optimizing the TDCA network using outer loop optimization. We demonstrated this using a three-layer fully connected feed-forward network, presenting the TDCA network's convergence process under the PGPE algorithm in Fig. \ref{PGPE_convergence}. The figure's x-axis shows the iteration steps, and the y-axis shows the cross-entropy and classification accuracy of the optimized bottom-up network.

We recorded the first 500 steps to observe convergence trends. Fig. \ref{PGPE_CE} and Fig. \ref{PGPE_acc} show the final cross-entropy loss and classification accuracy of the inner loop's bottom-up network, respectively. The TDCA network quickly converges and fine-tunes to optimal performance across various datasets, showing its effective optimization by the PGPE algorithm. All further bottom-up network results used the trained TDCA network.

\subsection*{Analysis of Top-down Training Process}

To understand the TDCA framework's dynamics, we studied the behavior of the TDCA-network and bottom-up networks during classification tasks. We focused on how the credit, generated by a trained TDCA-network, updates the bottom-up network. 

\subsubsection*{Convergence Analysis}

\begin{figure}[p]
	\graphicspath{{figures/analysis/}}
	\subfigure[MNIST]{
		\includegraphics[width=0.45\linewidth]{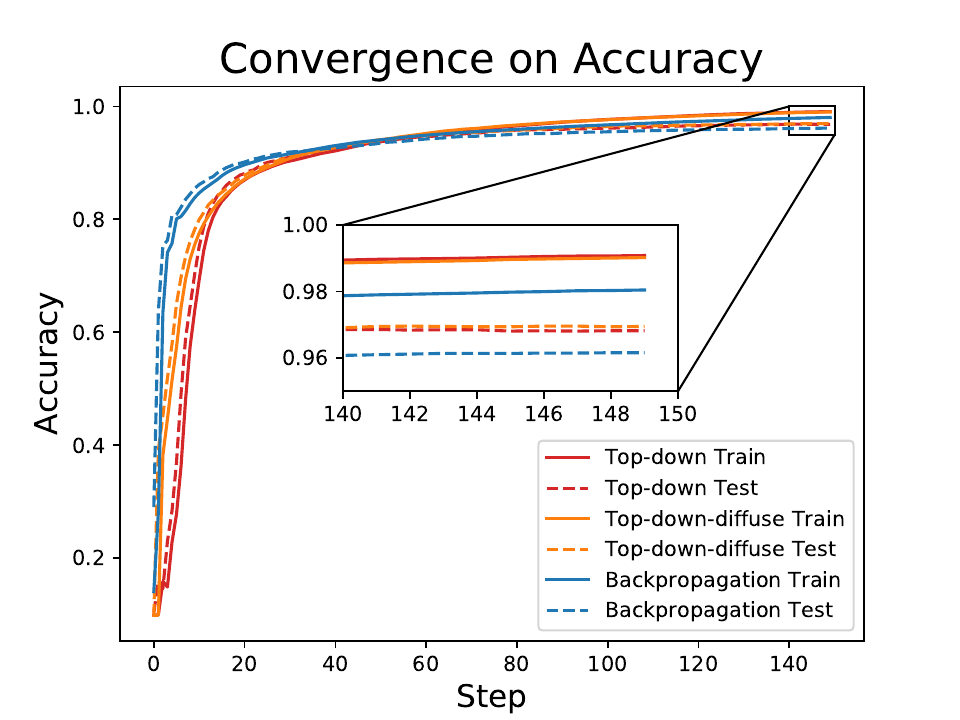}
	}
	\subfigure[Fashion-MNIST]{
		\includegraphics[width=0.45\linewidth]{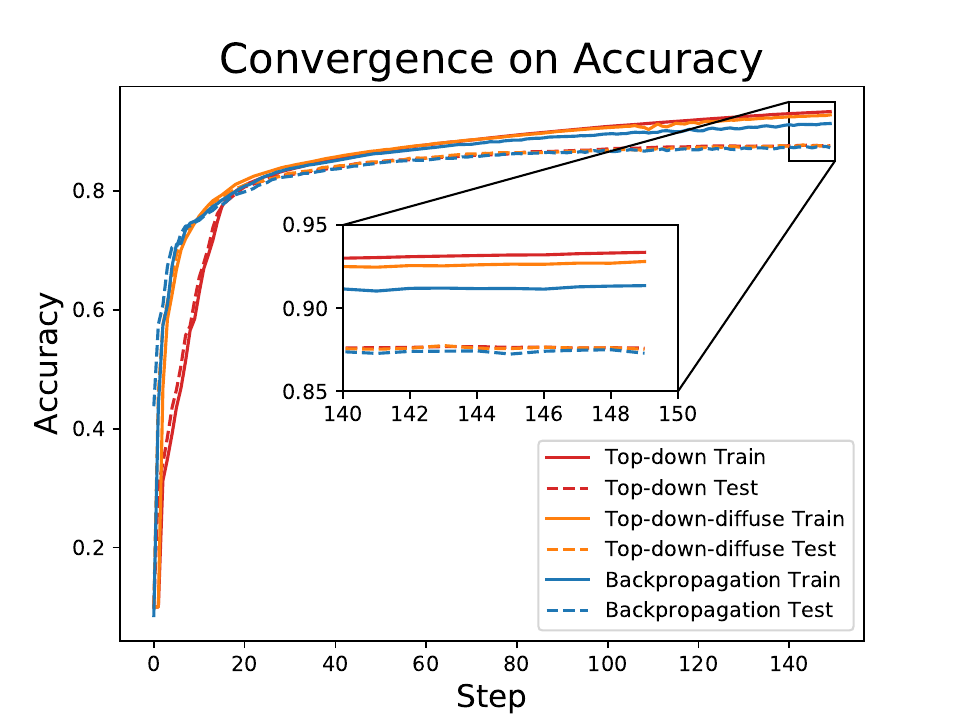}
	}
	
	\centering
	\caption{Convergence patterns of the bottom-up network across various datasets. The bottom-up network trained by back-propagation (blue line) exhibits rapid convergence but falls short in final performance. In contrast, the top-down line (red) converges more slowly but yields superior performance in the final, while the top-down diffuse line (orange) shows intermediate characteristics.}
	\label{convergence_curve}
\end{figure}

Fig. \ref{convergence_curve} shows the convergence curves of the bottom-up network. Our findings show that the bottom-up network, trained by traditional BP, converges faster than the TDCA-network initially (before 40 steps). This indicates that back-propagation quickly falls into the local optimum near the initialization. In contrast, the TDCA-network takes a long-term perspective, leading to superior performance on both the training and testing sets. These results show that the TDCA-network can bypass the local optimum early on, providing a more effective update strategy than BP.

\subsubsection*{Landscape Analysis}

\begin{figure}[p]
	\graphicspath{{figures/analysis/}}
	\subfigure[2D landscape (Dense vs. BP)]{
		\label{2d-td-bp}
		\includegraphics[width=0.45\linewidth]{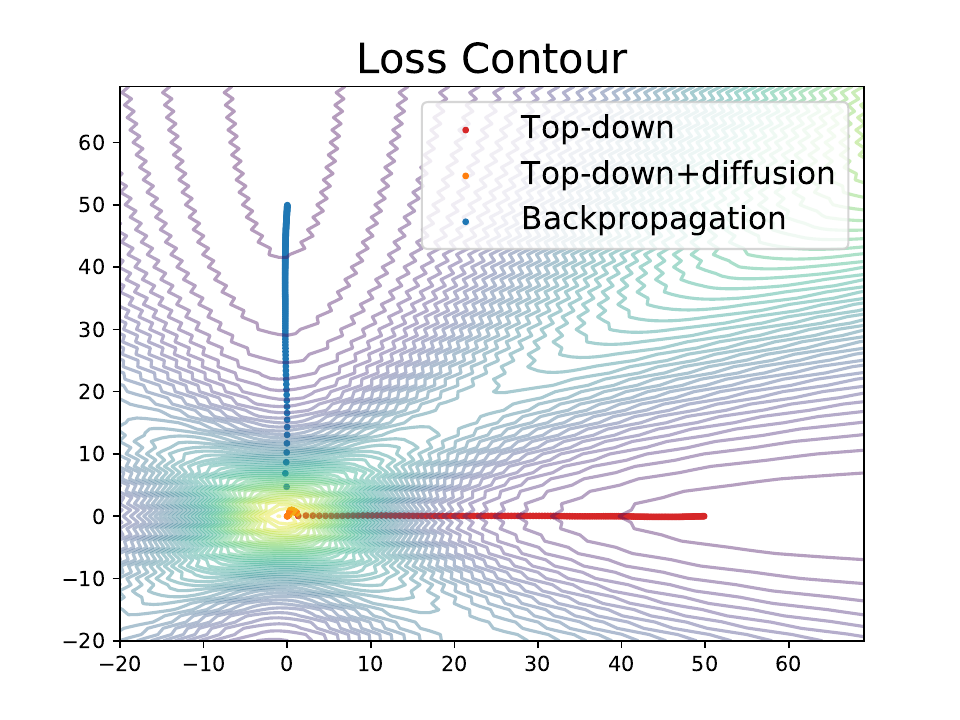}
	}
	\subfigure[3D landscape (Dense vs. BP)]{
		\label{3d-td-bp}
		\includegraphics[width=0.45\linewidth]{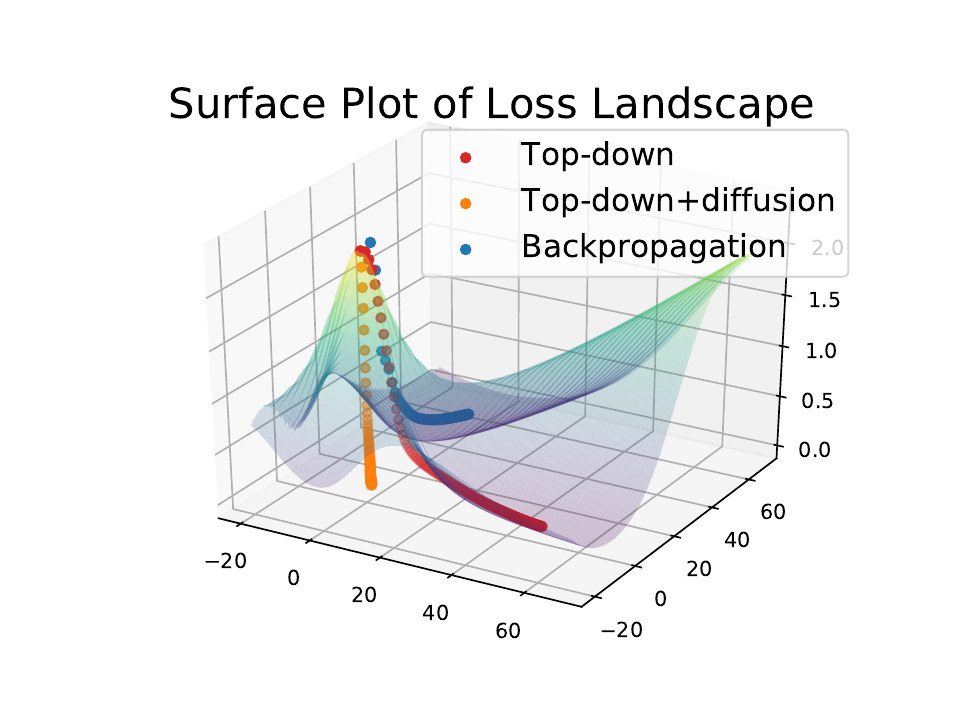}
	}

	\subfigure[2D landscape (Dense vs. Sparse)]{
		\label{2d-td-diffuse}
		\includegraphics[width=0.45\linewidth]{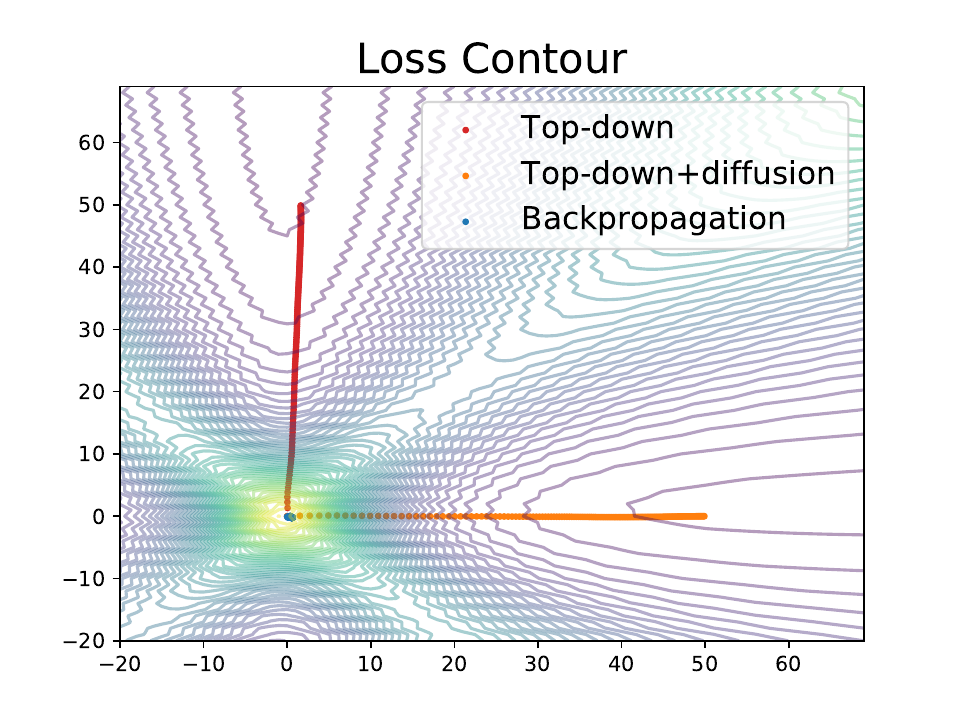}
	}
	\subfigure[3D landscape (Dense vs. Sparse)]{
		\label{3d-td-diffuse}
		\includegraphics[width=0.45\linewidth]{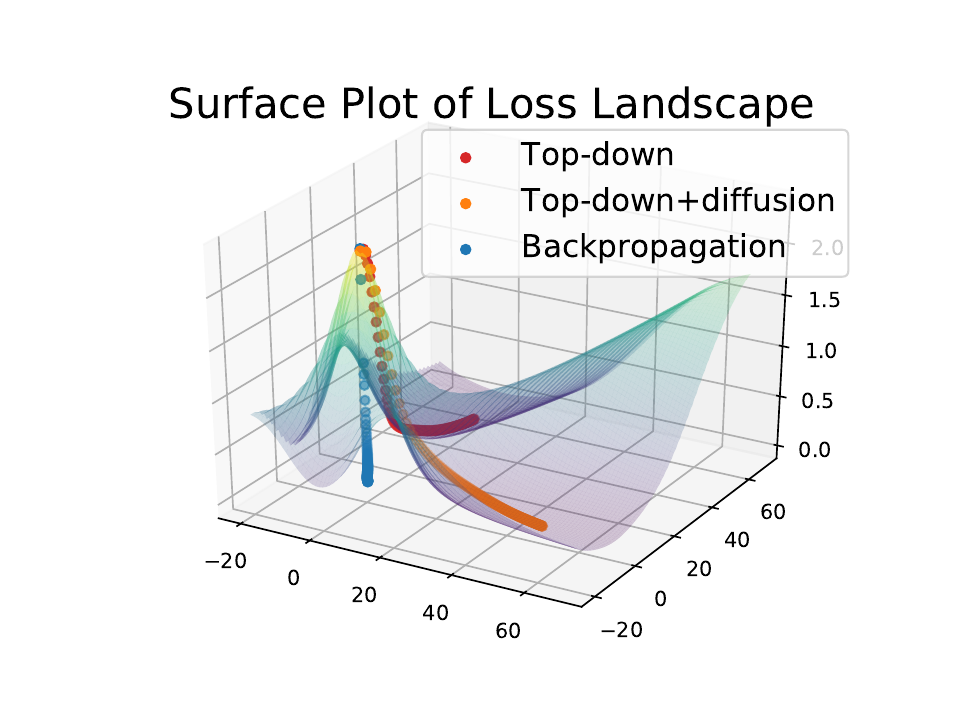}
	}
	
	\centering
	\caption{Parameter loss landscape of a three-layer bottom-up network on the MNIST dataset. Each landscape is defined by the update trajectory of two distinct algorithms. The update trajectories of Top-down (TD), Top-down+diffuse (Diff), and Backpropagation (BP) are almost orthogonal to each other in the TD-BP and TD-Diff spaces.}
	\label{landscape}
\end{figure}

To understand the behavior of the bottom-up network under different update strategies, we plotted the loss landscapes \cite{li2018visualizing} of the bottom-up networks and showed the trajectories of parameter updates under the supervision of the top-down method, top-down+diffusion method, and backpropagation (Fig. \ref{landscape}). When plotting the loss landscapes, we used the PCA direction of the top-down method and backpropagation (Fig. \ref{2d-td-bp} and Fig. \ref{3d-td-bp}), and the PCA direction of the top-down method and the top-down+diffusion method (Fig. \ref{2d-td-diffuse} and Fig. \ref{3d-td-diffuse}). We observed that the top-down trajectory and back-propagation trajectory are almost orthogonal (Fig. \ref{2d-td-bp}), as is the top-down+diffusion trajectory to the top-down trajectory (Fig. \ref{2d-td-diffuse}). Based on Fig. \ref{3d-td-bp} and Fig. \ref{3d-td-diffuse}, we concluded that the three methods generate distinct update trajectories for the bottom-up model, unrelated to each other.

The convergence analysis shows that the TDCA-network, within the top-down learning framework, can devise a more effective update strategy for the bottom-up network than BP. The orthogonal update trajectories of different methods highlight their unique optimization problem-solving strategies.


\clearpage

\clearpage
\begin{table*}
	\centering
	\caption{Classification performance on different datasets.}
	\resizebox{\linewidth}{!}{
		\renewcommand{\arraystretch}{1.25}
		\begin{tabular}{cccc}
			\hline
			Method &  MNIST  & Fashion-MNIST  & CIFAR-10  \\
			\hline
			Kaiming BP                  &  $98.20\pm0.15\% / 96.16\pm0.14\% $ & $91.26\pm0.38\% / 87.19\pm0.41\% $& $52.21\pm0.66\% / 38.99\pm0.51\%$ \\
			Zero BP                     &  $ 99.04\pm0.066\% / 96.35\pm0.15\%$ & $92.33\pm0.11\% / 87.48\pm0.12\%$ & $ 60.83\pm0.83\% / 40.42\pm0.41\% $ \\	
			TDCA-net                    &  99.08\% / 96.82\%  & 93.36\% / 87.59\% & 60.11\% / 42.03\% 	\\
			
			\hline

	\end{tabular}}
	\label{top-vs-bp-table}
\end{table*}

\begin{table*}
	\centering
	\caption{Generalization performance of TDCA-network  across datasets.}
	
	\resizebox{\linewidth}{!}{
		\renewcommand{\arraystretch}{1.25}
		\begin{tabular}{ccccc}
			\hline
			\diagbox{Evolve on} {Transfer to}                        & MNIST & Fashion-MNIST & CIFAR-10  & Mean \\
			\hline
			MNIST                          &  99.08\%/96.82\%  & 85.94\%/83.91\% & 41.476\%/36.27\% &  75.50\%/72.33\% \\
			
			Fashion-MNIST                    &  93.998\%/93.21\%  & 93.37\%/87.56\% & 40.85\%/35.67\%  & 76.07\%/72.15\% \\
			
			CIFAR-10                         &  90.48\%/89.78\%  & 84.47\%/82.8\% & 60.12\%/42.23\% & 78.35\%/71.60\% \\
			
			All                              & 96.22\%/94.92\% & 89.14\%/86.39\% & 59.35\%/42.25\% & 81.57\%/74.52\% \\
			\hline
			
	\end{tabular}}
	
	\label{transfer_table}
\end{table*}

\begin{table*}
	\centering
	\caption{Performance of TDCA-network with credit diffusion. }	\resizebox{\linewidth}{!}{
		\renewcommand{\arraystretch}{1.25}
		\begin{tabular}{cccccc}
			\hline
			Dataset &    100-to-100       & 10-to-100      & 100-to-1000    & 36-to-900  & 6*6-to-30*30  \\
			\hline
			MNIST             &  99.08\% / 96.82\%  & 99.20\% / 96.96\% &  99.67\% / 96.50\% & 98.01\% / 96.51\% & 99.87\% / 96.45\% \\
			
			Fashion-MNIST     &  93.37\% / 87.56\%  & 92.80\% / 87.51\%   & 92.81\% / 87.59\% & 34.87\% / 36.23\% &	93.80\% / 87.86\%\\
			
			CIFAR-10          & 60.12\% / 42.23\%   & 58.37\% / 41.33\% & 71.22\% / 39.42\%  & 66.09\% / 40.15\% & 73.74\% / 38.07\% \\
			\hline
			
	\end{tabular}}
	
	\label{diffuse_performance}
\end{table*}

\clearpage
\begin{table*}
	\centering
	\caption{Complexity of different methods.}	
	\resizebox{0.6\linewidth}{!}{
		\renewcommand{\arraystretch}{1.25}
		\begin{tabular}{lcccc}
			\hline
			\multirow{2}{*}{Settings}              & \multicolumn{2}{c}{MNIST/FashionMNIST} & \multicolumn{2}{c}{CIFAR-10} \\ 
			\cmidrule(lr){2-3}  \cmidrule(lr){4-5}
			& Param               & MFLOPs           & Param          & MFLOPs       \\
			\hline
			BP(100 hidden)       & 1000                & 240              & 1000         & 200          \\
			Credit parameters   & 4892794             & \mbox{-}         & 18807870       & \mbox{-}     \\
			Credit neurons      & 4970                & 576              & 4970           & 480          \\
			Credit groups      & 2180                & 252              & 2180           & 210   	\\
			\hline 
			BP(900hidden)      &  9000         & 2160             &  9000          & 1800    \\
			Credit groups       & 2986       & 345.6            &  2986          & 288       \\
			\hline
		\end{tabular}
	}
	
	\label{complexity}
\end{table*}

\begin{table}
	\centering
	\renewcommand{\arraystretch}{1.25}
	\caption{Comparison of TDCA-network and other algorithms on optimizing reinforcement tasks.}
	\label{reinforce-table}
	\resizebox{0.8\linewidth}{!}{
		\begin{tabular}{lcccc}
			\hline
			Method &  CartPole-v1  & Pendulum-v0 & BipedalWalker-v3  \\
			\hline
			Reward Range                    & [0,200] & [-3254.7,0] & [-400, 300] \\
			Policy Gradient           &  199 (success) &  -  & -    \\
			TD3                    &  -  &  -647 (success) &   -88 (False)    	\\
			TDCA-net            & 200 (success) & -119 (success) & 288.4 (success) \\
			\hline

		\end{tabular}
	}
\end{table}

\begin{table}
	\centering
	\renewcommand{\arraystretch}{1.25}
	\caption{Multi agents controlled by the same TDCA-network.}
	\label{reinforce-multi-task-v2-parallel-table}
	\setlength{\tabcolsep}{2mm}
	\resizebox{\linewidth}{!}{
		\begin{tabular}{ccccc}
			\hline
			Measurement &  Button-press-wall-v2  & Dial-turn-v2 & Door-close-v2  \\
			\hline
			Training (3 init)                          &  2581.9 (success 0\%) &  1497.5 (success 100\%)  & 3451.2 (success 100\%)    \\
			Test (10 init)                    &  1932.2 (success 0\%)  &  1493.2 (success 80\%) &   3385.5 (success 90\%)    	\\
			\hline
			
	\end{tabular}}
\end{table}

\end{document}